\documentclass[sigconf]{acmart}

\AtBeginDocument{%
  \providecommand\BibTeX{{%
    \normalfont B\kern-0.5em{\scshape i\kern-0.25em b}\kern-0.8em\TeX}}}
    
\usepackage{graphicx}
\usepackage{subcaption}
\usepackage{wrapfig}

\setcopyright{acmcopyright}
\copyrightyear{2019}
\acmYear{2019}
\acmDOI{_}

\acmConference[KDD-ADF '19]{2nd KDD Workshop on Anomaly Detection in Finance}{August 05, 2019}{Anchorage, Alaska}
\acmBooktitle{}
\acmPrice{15.00}
\acmISBN{}

\begin{document}

\title{Detection of Accounting Anomalies in the Latent Space using Adversarial Autoencoder Neural Networks}

\author{Marco Schreyer}
\affiliation{%
    \institution{University of St. Gallen}
    \city{St. Gallen}
    \country{Switzerland}
}
\email{marco.schreyer@unisg.ch}

\author{Timur Sattarov}
\affiliation{%
    \institution{Deutsche Bundesbank}
    \city{Frankfurt am Main}
    \country{Germany}
}
\email{timur.sattarov@bundesbank.de}

\author{Christian Schulze}
\affiliation{%
    \institution{University of St. Gallen}
    \city{St. Gallen}
    \country{Switzerland}
}
\email{christian.schulze@unisg.ch}

\author{Bernd Reimer}
\affiliation{%
    \institution{PricewaterhouseCoopers GmbH WPG}
    \city{Stuttgart}
    \country{Germany}
}
\email{reimer.bernd@pwc.com}

\author{Damian Borth}
\affiliation{%
    \institution{University of St. Gallen}
    \city{St. Gallen}
    \country{Switzerland}
}
\email{damian.borth@unisg.ch}


\renewcommand{\shortauthors}{Schreyer and Sattarov, et al.}

\begin{abstract}
The detection of fraud in accounting data is a long-standing challenge in financial statement audits. Nowadays, the majority of applied techniques refer to handcrafted rules derived from known fraud scenarios. While fairly successful, these rules exhibit the drawback that they often fail to generalize beyond known fraud scenarios and fraudsters gradually find ways to circumvent them. In contrast, more advanced approaches inspired by the recent success of deep learning often lack seamless interpretability of the detected results. To overcome this challenge, we propose the application of adversarial autoencoder networks. We demonstrate that such artificial neural networks are capable of learning a semantic meaningful representation of real-world journal entries. The learned representation provides a holistic view on a given set of journal entries and significantly improves the interpretability of detected accounting anomalies. We show that such a representation combined with the networks reconstruction error can be utilized as an unsupervised and highly adaptive anomaly assessment. Experiments on two datasets and initial feedback received by forensic accountants underpinned the effectiveness of the approach.
\end{abstract}



\maketitle

\section{Introduction}
\label{sec:introduction}

The Association of Certified Fraud Examiners estimates in its "Global Study on Occupational Fraud and Abuse 2018" \cite{ACFE2018} that organizations lose 5\% of their annual revenues due to fraud. The term "fraud" refers to "the abuse of one's occupation for personal enrichment through the deliberate misuse of an organization's resources or assets" \cite{Wells2017}. A similar study, conducted by the auditors of PwC, revealed that approx. 30\% of the respondents experienced losses between \$100,000 USD and \$5 million USD due to fraud \cite{PWC2018}. The study also showed that financial statement fraud caused by far the highest median loss of the surveyed fraud schemes\footnote{The ACFE study encompasses an analysis of 2.690 cases of occupational fraud investigated between January 2016 and October 2017 that occurred in 125 countries. The PwC study encompasses over 7.228 respondents that experienced economic crime in the last 24 months.}. 

\begin{figure}[t!]
	\hspace*{0.0cm} \includegraphics[width=9cm, angle=0, trim={1.6cm 2.0cm 0.0 0.0}]{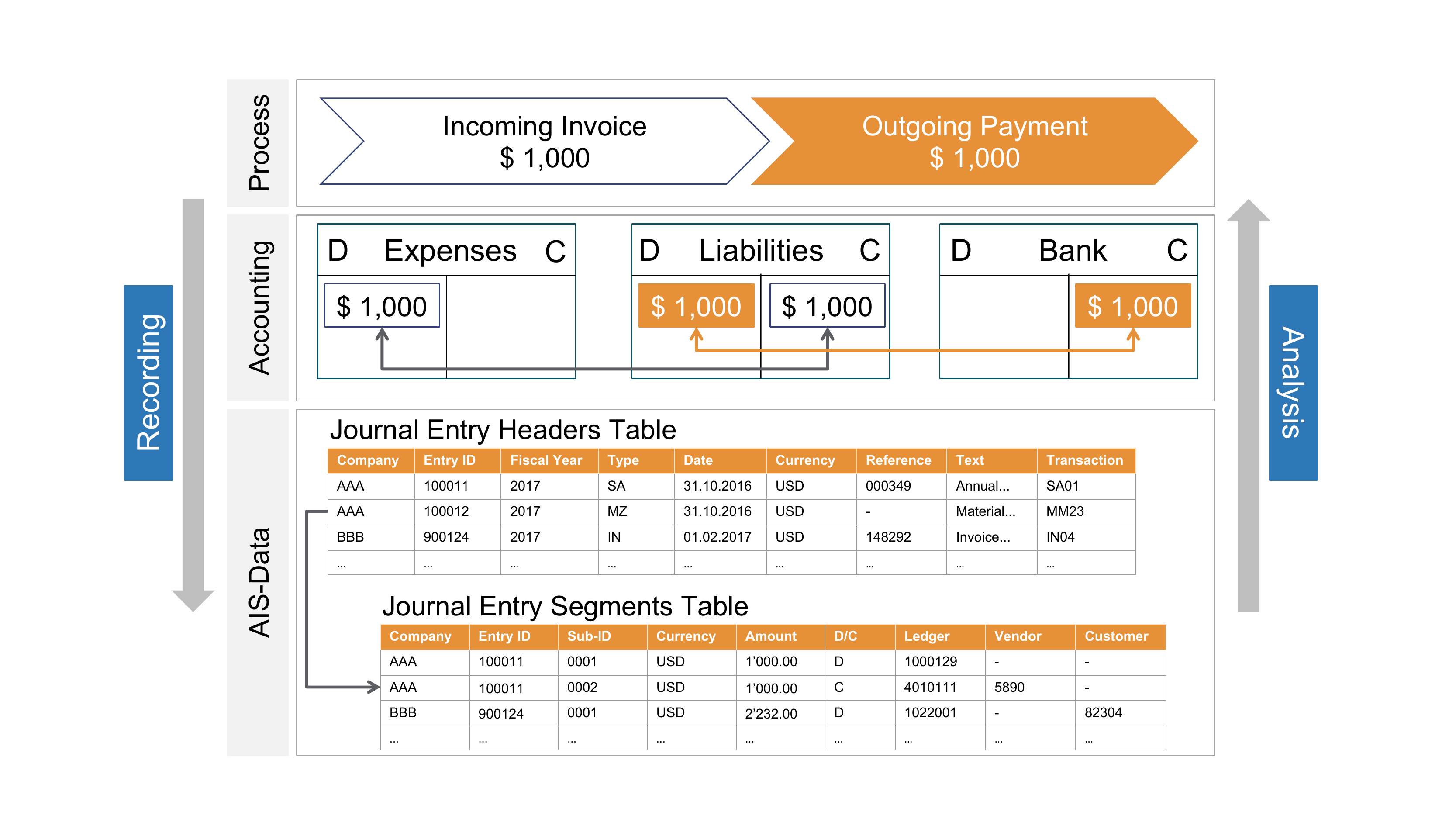}
	\caption{Hierarchical view of an Accounting Information System (AIS) that records distinct layer of abstractions, namely (1) the business process, (2) the accounting and (3) technical journal entry information in designated tables.}
	\label{fig:ais_system}
\end{figure}

At the same time, organizations accelerate the digitization of business processes \cite{McKinsey2014} affecting in particular Accounting Information Systems (AIS) or more generally Enterprise Resource Planning (ERP) systems. Steadily, these systems collect vast quantities of business process and accounting data at a granular level. This holds in particular for the journal entries of an organization recorded in its general ledger and sub-ledger accounts. SAP, one of the most prominent enterprise software providers, estimates that approx. 77\% of the world's transaction revenue touches one of their ERP systems \cite{SAP2019}. Figure \ref{fig:ais_system} depicts a hierarchical view of an AIS recording process of journal entry information in designated database tables.

\begin{figure*}[ht!]
    \center
    \includegraphics[height=6.5cm]{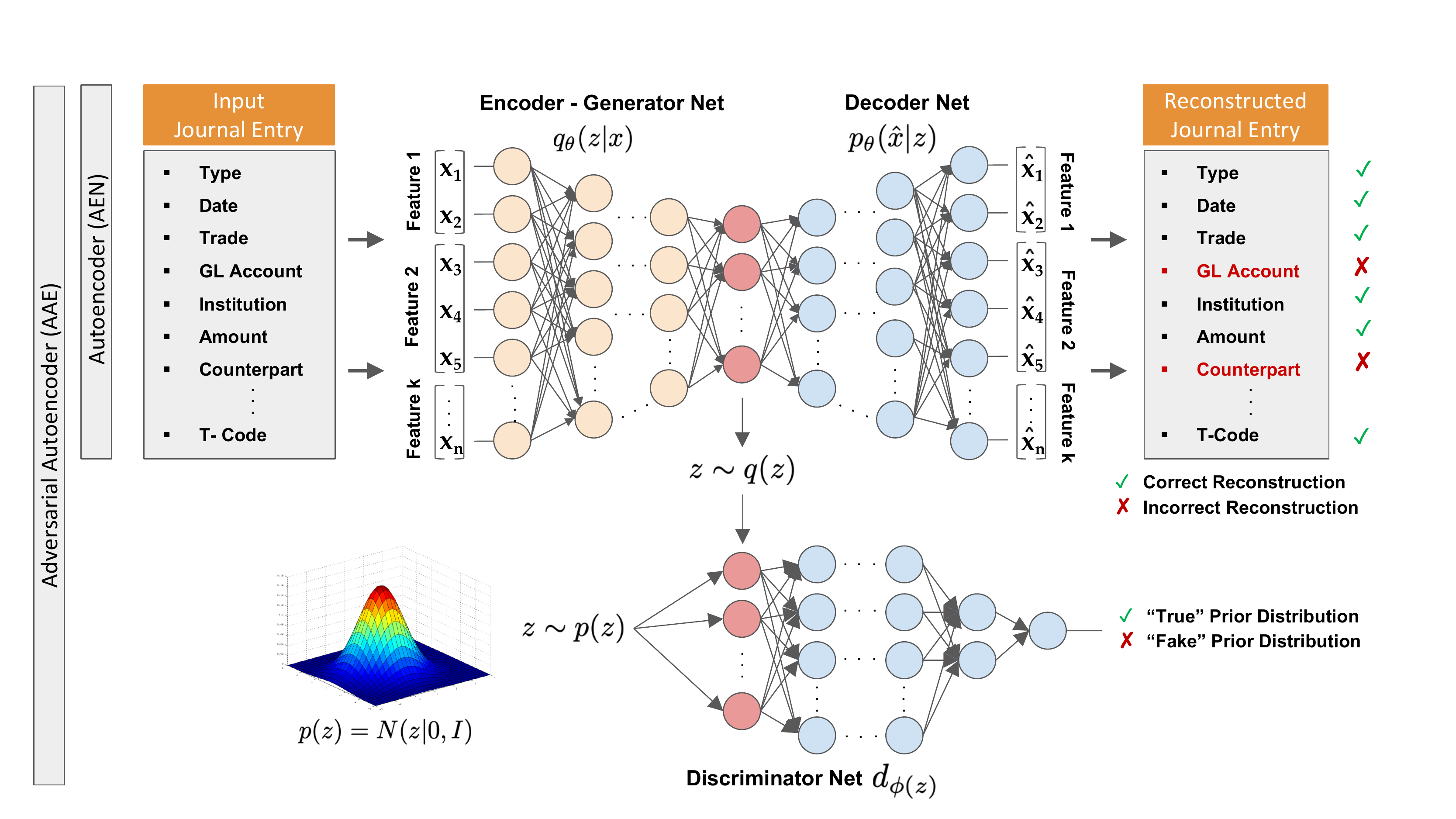}
    \caption{The adversarial autoencoder architecture \cite{makhzani2015}, applied to learn the journal entries characteristics and to partition the entries into semantic meaningful groups. The adversarial autoencoder network architecture imposes an arbitrary prior distribution $p(z)$ on the discrete latent code vector $z$, e.g., a mixture of Gaussians. With progressing training the encoder learns an aggregated posterior distribution $g_{\theta}(z|x)$ that matches the imposed prior to fool the discriminator network $d_{\phi}$.}
    \label{fig:architecture}
\end{figure*}

In order to conduct fraud, perpetrators need to deviate from the regular system usage or posting pattern. Such deviations are recorded by a very limited number of "anomalous" journal entries and their respective attribute values. To detect potentially fraudulent activities international audit standards require the direct assessment of such journal entries \cite{AICPA2002}, \cite{IFAC2009fraud}. Nowadays, auditors and forensic accountants apply a wide range of data analysis techniques to examine journal entries during an audit. These techniques often encompass rule-based analyses referred to as "red-flag" tests (e.g. postings late at night, multiple vendor bank account changes) as well as statistical analyses (e.g. Benford's Law, time series evaluation). Nevertheless, the detection of traces of fraud in up to several hundred million journal entries remains a labor-intensive task requiring significant time effort and resources. 

Driven by the recent technological advances in artificial intelligence \cite{LeCun2015} deep neural network-based techniques (e.g. deep autoencoder neural networks) have emerged into the field of forensic accounting and financial statement audits \cite{schreyer2017}. Such, approaches often lack a seamless interpretability of the detected "anomalous" journal entries selected for a detailed audit. This is a major drawback since auditors are required to test a representative sample of journal entries in order to reduce the "sampling risk" \cite{IFAC2009sampling} of an audit. Ultimately, the testing of an individual entry while ignoring another one must be defensible in court \cite{hall2002}.

To overcome this challenge we propose the application of Adversarial Autoencoder Neural Networks (AAEs) \cite{makhzani2015}. We demonstrate that such adversarial architectures are capable of learning a semantic meaningful representation of journal entries. The learned representation allows for an improved interpretability of the entries underlying generative processes as well as detected "anomalous" journal entries. In summary, we present the following contributions: 

\begin{itemize}

\item We illustrate that AAEs can be used to learn a representation of journal entries observable in real-world ERP systems that partitions the entries into semantic meaningful groups;

\item We demonstrate how such a learned representation can be used by a human auditor or forensic accountant to sample journal entries for an audit in an interpretable manner;

\item We show that the learned representation combined with the magnitude of an entry's reconstruction error can be interpreted as a highly adaptive anomaly assessment of journal entries.

\end{itemize}

We envision this deep learning-based methodology as an important supplement to the auditors and forensic accountants toolbox \cite{Pedrosa2014}. The remainder of this work is structured as follows: In Section \ref{sec:relatedwork} we provide an overview of the related work. Section \ref{sec:methodology} follows with a description of the adversarial autoencoder network architecture and presents the proposed methodology to detect accounting anomalies. The experimental setup and results are outlined in Section \ref{sec:experiments} and Section \ref{sec:results}. In Section \ref{sec:conclusion} the paper concludes with a summary of the current work and future directions of research. An implementation of the proposed methodology is available at \url{https://github.com/GitiHubi/deepAD}.

\section{Related Work}
\label{sec:relatedwork}

The literature survey presented hereafter focuses on (1) the detection of fraudulent activities in Enterprise Resource Planning (ERP) data and (2) the detection of financial fraud using deep Autoencoder Neural Networks (AENs) \cite{hinton2006} as well as Generative Adversarial Networks (GANs) \cite{Goodfellow2014}.

\subsection{Fraud Detection in Accounting Data}

The task of detecting fraud and accounting anomalies has been studied both by practitioners \cite{Wells2017} and academia \cite{Amani2017}. Several references describe different fraud schemes and ways to detect unusual and "creative" accounting practices \cite{Singleton2006}.

The forensic analysis of journal entries emerged with the advent of Enterprise Resource Planning (ERP) systems and the increased volume of data recorded by such systems. Bay et al. in \cite{Bay2002} use Naive Bayes methods to identify suspicious general ledger accounts, by evaluating attributes derived from journal entries measuring any unusual general ledger account activity. Their approach is enhanced by McGlohon et al. applying link analysis to identify (sub-) groups of high-risk general ledger accounts \cite{McGlohon2009}. Kahn et al. in \cite{Khan2009} and \cite{Khan2010} create transaction profiles of SAP ERP users. The profiles are derived from journal entry based user activity pattern recorded in two SAP R/3 ERP system in order to detect suspicious user behavior and segregation of duties violations. Similarly, Islam et al. in \cite{Islam2010} use SAP R/3 system audit logs to detect known fraud scenarios and collusion fraud via a "red-flag" based matching of fraud scenarios. Debreceny and Gray in \cite{Debreceny2010} analyze dollar amounts of journal entries obtained from 29 US organizations. In their work, they search for violations of Benford's Law \cite{Benford1938}, anomalous digit combinations as well as an unusual temporal pattern such as end-of-year postings. More recently, Poh-Sun et al. in \cite{Seow2016} demonstrate the generalization of the approach by applying it to journal entries obtained from 12 non-US organizations. Jans et al. in \cite{Jans2010} use latent class clustering to conduct a uni- and multivariate clustering of SAP ERP purchase order transactions. Transactions significantly deviating from the cluster centroids are flagged as anomalous and are proposed for a detailed review by auditors. The approach is enhanced in \cite{Jans2011} by a means of process mining to detect deviating process flows in an organization procure to pay process. Argyrou et al. in \cite{Argyrou2012} evaluate self-organizing maps to identify "suspicious" journal entries of a shipping company. In their work, they calculated the Euclidean distance of a journal entry and the code-vector of a self-organizing maps best matching unit. In subsequent work, they estimate optimal sampling thresholds of journal entry attributes derived from extreme value theory \cite{Argyrou2013}.

Concluding from the reviewed literature, the majority of references either draw from accounting and forensic knowledge about historical fraud schemes or non deep-learning based techniques to detect financial fraud. However, driven by the recent success of deep learning techniques, which are potentially misused by fraudsters, we see a high demand for auditors to likewise enhance their examination methodologies.    

\subsection{Anomaly Detection using Deep Learning}

Nowadays, deep learning inspired methods are increasingly used for novelty and anomaly detection in financial data \cite{chalapathy2019, pimentel2014}.

Renstr{\"o}m and Holmsten in \cite{renstrom2018} evaluate AENs to detect fraud in credit card transactions. Similarly, Kazemi and Zarrabi \cite{kazemi2017} and Sweers et al. \cite{sweers2018} train and evaluate a variety of variational AEN architectures. Pumsirirat and Yan in \cite{pumsirirat2018} compare the anomaly detection performance of AENs based on three datasets of credit card transactions. Wedge et al. \cite{wedge2017} use AENs to learn behavioral features from historical credit card transactions. Paula et al. in \cite{Paula2017} use AENs in export controls to detect traces of money laundry and fraud by analyzing volumes of exported goods. Similarly, Schreyer et al. in \cite{schreyer2017} utilized the reconstruction error of deep AENs to detect anomalous journal entries in two datasets of real-world accounting data. 

More recently, GANs are utilized in the context of fraud detection. Fiore et al. in \cite{fiore2017} train such networks to generate mimicked anomalies, which were used to augment training data to improve credit card fraud detection classifiers. Choi et al. in \cite{choi2018} train ensembles of generative models to successfully detect anomalies in credit card transactions. Zheng et al. in \cite{zheng2018b} train LSTM-AENs in an adversarial training set up to detect fraudulent credit card transactions. In another study, Zheng et al. in \cite{zheng2018a} propose generative denoising GANs to detect telecommunication fraud in the transactions of two financial institutions.

To the best of our knowledge, this work presents the first deep-learning inspired methodology trained in an adversarial training setup to detect anomalous journal entries in real-world accounting data.

\section{Methodology}
\label{sec:methodology}

To detect anomalous journal entries one first has to define "normality" with respect to accounting data. We assume that the majority of journal entries recorded within an organizations' ERP system relate to regular day-to-day business activities. In order to conduct fraud, perpetrators need to deviate from the "normal". Such a deviating behavior will be recorded by a very limited number of journal entries and their respective attribute values. We refer to journal entries exhibiting such deviating attribute values as \textit{accounting anomalies}.

\subsection{Accounting Anomaly Classes}

When conducting a detailed examination of real-world journal entries, recorded in large-scaled ERP systems, two characteristics can be observed: First, journal entry attributes exhibit a high variety of distinct attribute values, e.g., due to the high number of vendors or distinct posting amounts, and second, journal entries exhibit strong dependencies between certain attribute values e.g. a document type that is usually posted in combination with a certain general ledger account. Derived from this observation and similarly to Breunig et al. in \cite{Breunig2000} we distinguish two classes of anomalous journal entries, namely \textit{global} and \textit{local anomalies}:

\textbf{Global accounting anomalies} are journal entries that exhibit unusual or rare individual attribute values. Such anomalies usually relate to skewed attributes, e.g., rarely used ledgers, or unusual posting times. Traditionally, "red-flag" tests performed by auditors during an annual audit, are designed to capture this type of anomaly. However, such tests often result in a high volume of false-positive alerts due to rare but regular events such as reverse postings, provisions and year-end adjustments usually associated with a low fraud risk \cite{schreyer2017}. Furthermore, when consulting with auditors and forensic accountants, "global" anomalies often refer to "error" rather than "fraud". 

\textbf{Local accounting anomalies} are journal entries that exhibit an unusual or rare combination of attribute values while their individual attribute values occur quite frequently, e.g., unusual combinations of general ledger accounts or user accounts used by several accounting departments. This type of anomaly is significantly more difficult to detect since perpetrators intend to disguise their activities by imitating a regular activity pattern. As a result, such anomalies usually pose a high fraud risk since they correspond to processes and activities that may not be conducted in compliance with organizational standards.

We aim to learn a model that detects both classes of anomalous journal entries in an unsupervised manner. Thereby, the learned model should partitions the population of journal entries into semantic meaningful classes that allows for an increased interpretability of the detection results. To achieve this two-fold objective we utilize Adversarial Autoencoders (AAEs), a deep neural network architecture introduced by Makhzani et al. \cite{makhzani2015}. We provide preliminaries of Autoencoder Neural Networks (AENs) and Generative Adversarial Networks (GANs) that constitute the AAEs in the following. A more detailed presentation can be found in \cite{goodfellow2016}.

\subsection{Autoencoder Neural Networks}

Formally, let $X$ denote a set of $N$ journal entries $x^{1}, x^{2}, ..., x^{n}$, where each journal entry $x^{i}$ consists of $K$ attributes $x_{1}^{i}, x_{2}^{i}, ... , x_{j}^{i}, ... ,x_{k}^{i}$. Thereby, $x_{j}^{i}$ denotes the $j$-th attribute of the $i$-th journal entry. The individual attributes $x_{j}$ describe the journal entries accounting specific details, e.g., the entries fiscal year, posting type, posting date, amount, general-ledger. Hinton and Salakhutdinov in \cite{hinton2006} introduced AENs, a special type of feed-forward multi-layer network that can be trained to reconstruct its input. Formally, AENs are comprised of two nonlinear functions referred to as encoder $q_\theta$ and decoder $p_\theta$ network \cite{rumelhart1985}. The encoder function $q_\theta(\cdot)$ maps the input $x \in \mathcal{R}^k$ to a code vector $Z \in \mathcal{R}^m$ referred to as \textit{latent space representation}, where usually $k > m$. This latent representation is then mapped back by the decoder function $p_\theta(\cdot)$ to a reconstruction $\hat{x} \in \mathcal{R}^k$ of the original input space. In an attempt to achieve $x \approx \hat{x}$ the AEN is trained to minimize the dissimilarity of a given journal entry $x^{i}$ and its reconstruction $\hat{x}^{i} = p_\theta(q_\theta(x^{i}))$ as faithfully as possible. Thereby, the training objective is to learn a set of optimal model parameters $\theta^{*}$ by minimizing the AENs reconstruction loss, formally denoted as:

\begin{equation}
    \arg \min_{\theta} \|x^{i} - p_{\theta}(q_{\theta}(x^{i}))\|.
    \label{equ:reconstruction_loss}
\end{equation}

\subsection{Generative Adversarial Neural Networks}

Goodfellow et al. introduced GANs in \cite{goodfellow2014b}, a framework for training deep generative models using a mini-max game. The objective is to learn a generator distribution $q(x)$ that matches the real data distribution $p_{d}(x)$ of journal entries. Instead of trying to explicitly assign probability to every $x^{i}$ in the data distribution, the GAN aims to learn a set of parameters $\theta$ of a generator network $q_{\theta}$ that generates samples from the generator distribution $q(x)$ by transforming a noise variable $z \sim p_{n}(z)$ into a sample $q_{\theta}(z)$. Thereby, the generator is trained by playing against an adversarial discriminator network $d_{\phi}$ that aims to learn a set of parameters $\phi$ to distinguish between samples from the true data distribution $p_{d}$ and the generator's distribution $q(z)$. Both networks establish a min-max adversarial game. A solution to this game can be, expressed as:

\begin{equation}
    \min_{q_{\theta}} \max_{d_{\phi}} \mathbb{E}_{x \sim {p_{d}(x^{i})}}[\log d_{\phi}(x^{i})] + \mathbb{E}_{z\sim{p_{n}(z^{i})}}[\log(1-d_{\phi}(q_{\theta}(z^{i}))].
    \label{equ:adversarial_loss}
\end{equation}

\subsection{Adversarial Autoencoders}

The AAE architecture, as illustrated in Fig. \ref{fig:architecture}, extends the concept of AEN by imposing an arbitrary prior on the AENs latent space using a GAN training setup \cite{makhzani2015}. This is achieved by training the AAE jointly in two phases (1) a reconstruction phase as well as (2) an adversarial regularization phase.

In the reconstruction phase, the AAEs encoder network $q_{\theta}(z|x)$ is trained to learn an aggregated posterior distribution $q(z)$ of the journal entries $X$ over the latent code vector $Z$. Thereby, the learned posterior distribution corresponds to a compressed representation of the journal entry characteristics. Similarly to AENs, the decoder network $p_{\theta}(\hat{x}|z)$ of the AAE utilizes the learned latent code vector representations $Z$ to reconstruct the journal entries $\hat{X}$ as faithfully as possible to minimize the AAEs reconstruction error.

In the regularization phase, an adversarial training setup is applied were the encoder network $q_{\theta}(z|x)$ of the AAE functions as the generator network. In addition, a discriminator network $d_{\phi}(z)$ is attached on top of the learned latent code vector $Z$. Similarly to GANs, the discriminator network of the AAE is trained to distinguish samples of an imposed prior distribution $p(z)$ onto $Z$ from the learned aggregated posterior distribution $q(z)$. In contrast, the encoder network is trained to learn a posterior distribution $p(z) \approx q(z)$ that fools the discriminator network into thinking that the samples drawn from $q(z)$ originate from the imposed prior distribution $p(z)$.

\subsection{Accounting Anomaly Detection}

\begin{figure*}[ht!]
    \begin{center}
        \includegraphics[width=0.3\textwidth]{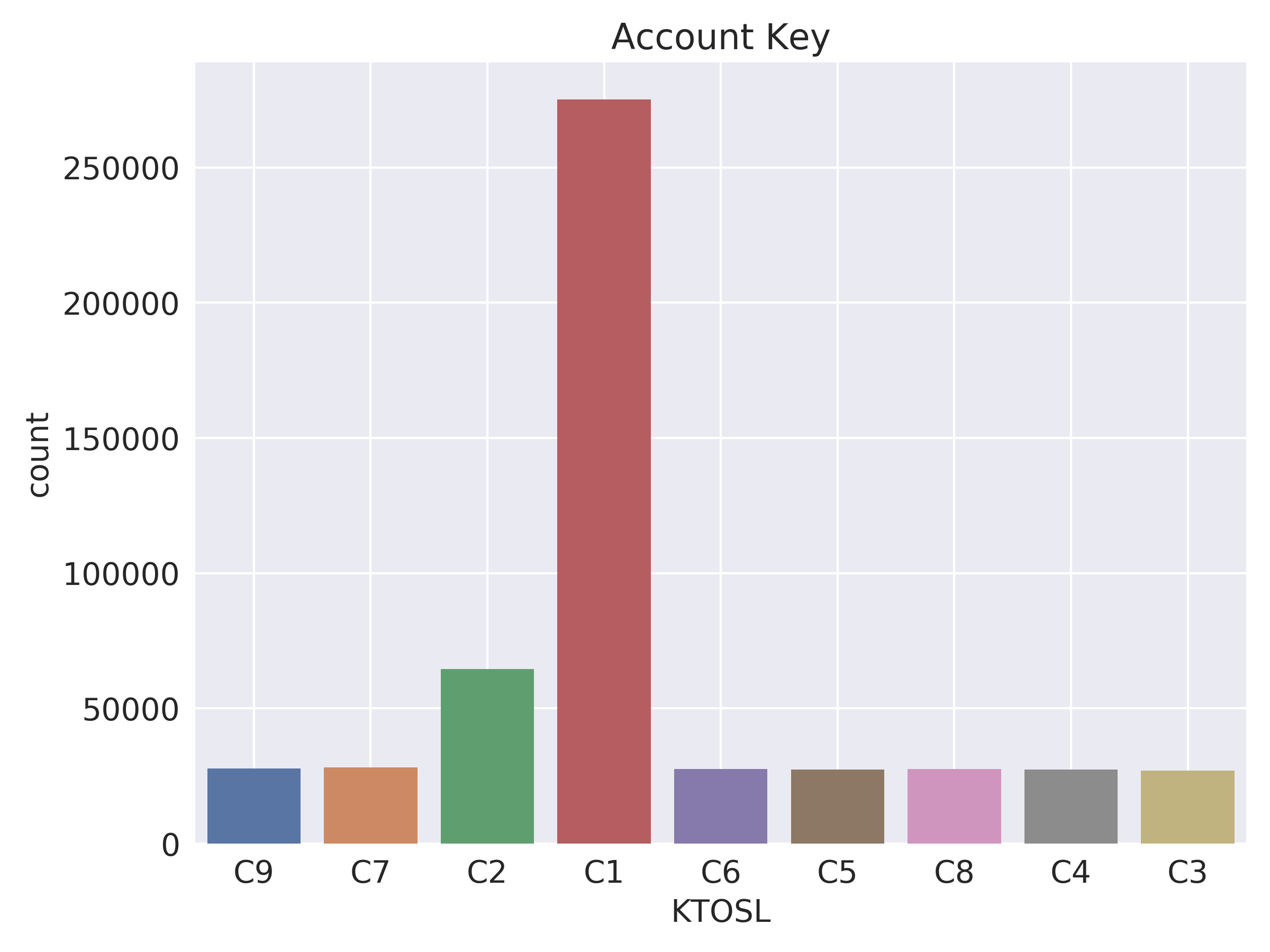}
        \includegraphics[width=0.3\textwidth]{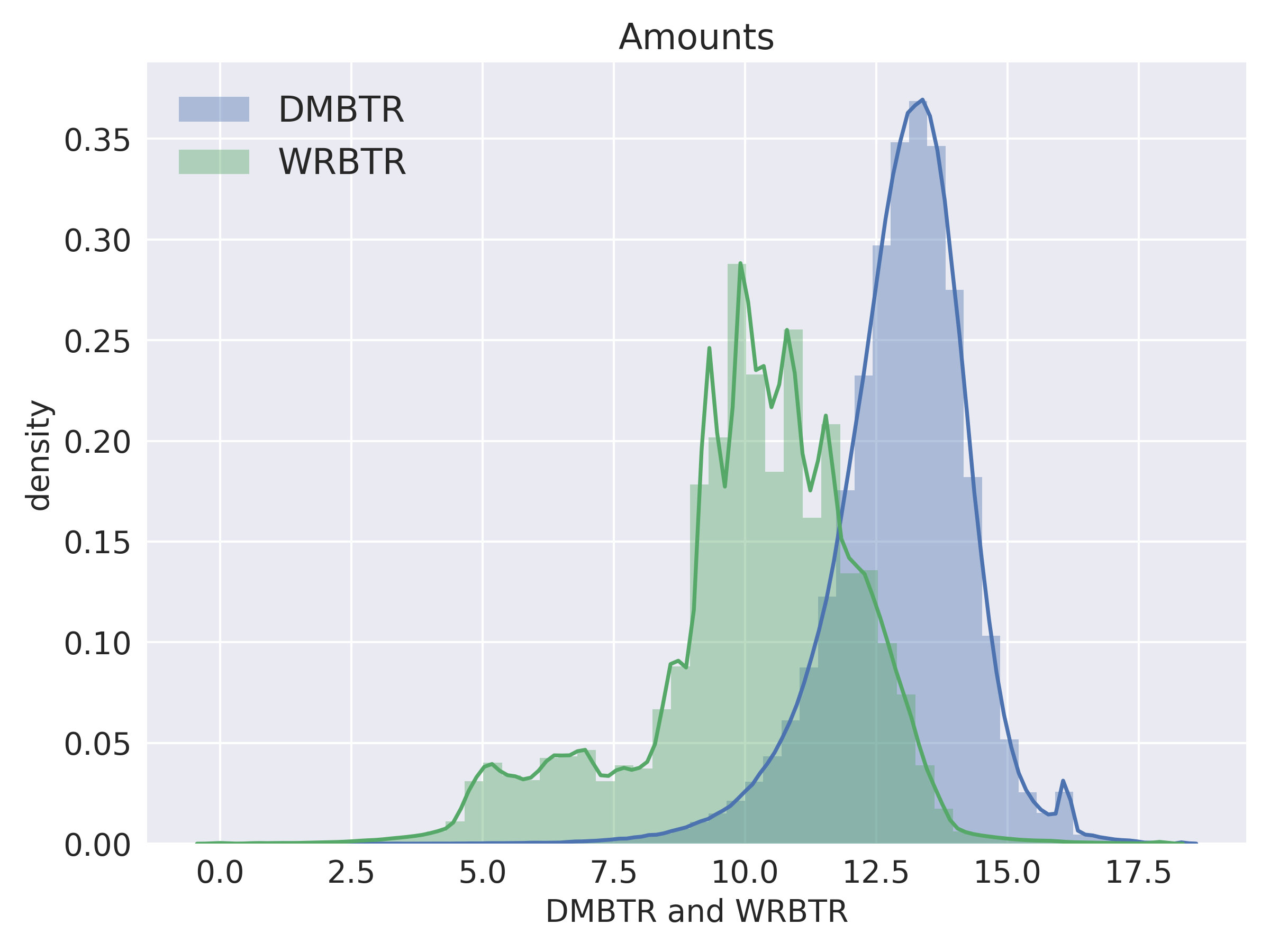}
         \includegraphics[width=0.3\textwidth]{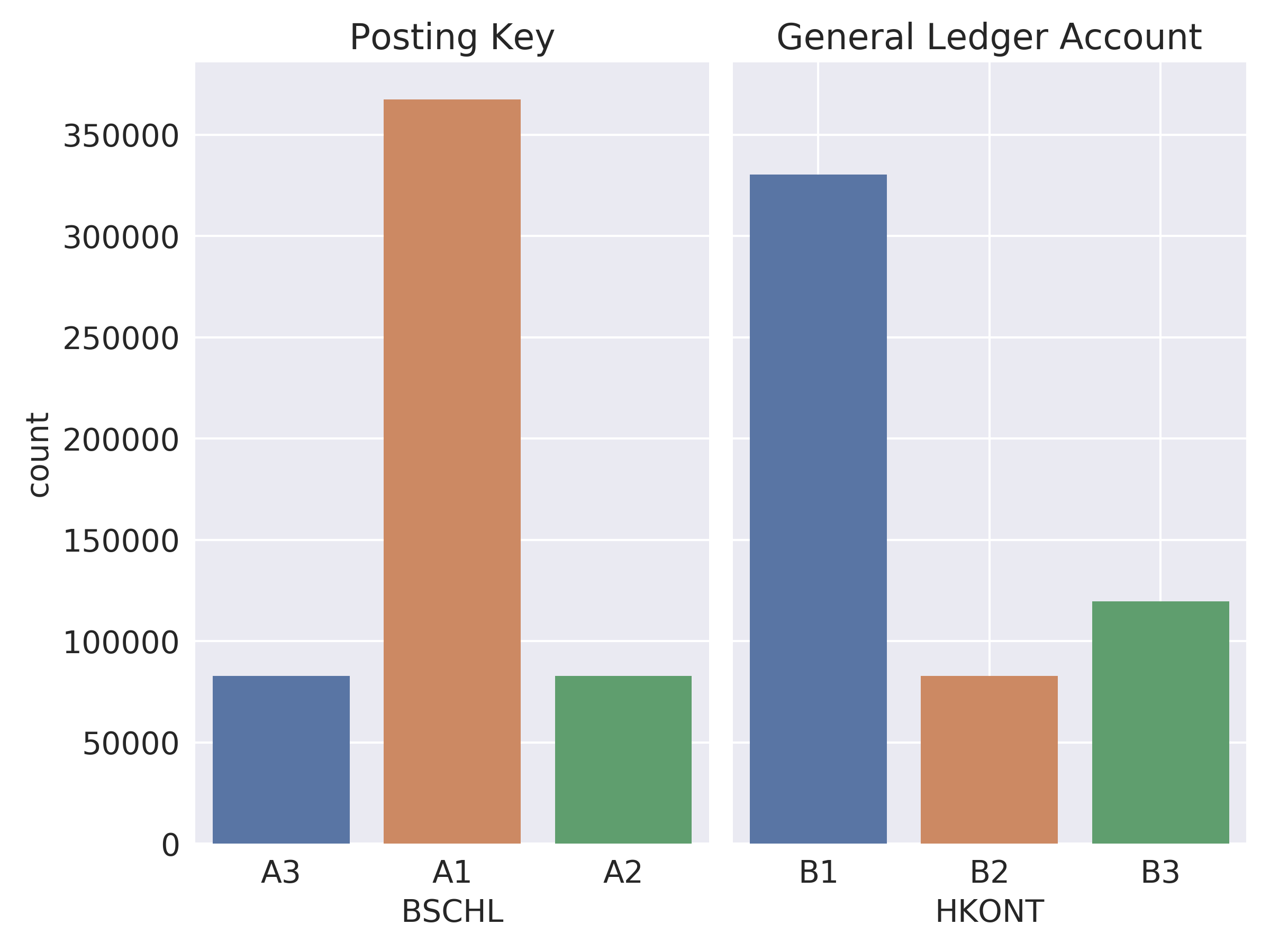}
    \end{center}
    \caption{Exemplary distribution of the 'account key' (technically: 'KTOSL') attribute values (left), the log-normalized 'local' and 'foreign currency amount' (technically: 'DMBTR' and 'WRBTR') attribute values (center), as well as the 'posting key' (technically: 'BSCHL') and 'general ledger account' (technically: 'HKONT') (right) observable in Dataset B.}
    \label{fig:input_data_distribution}
\end{figure*}

In order, to detect interpretable accounting anomalies in real-world ERP datasets we propose a novel anomaly score utilizing the introduced AAE architecture. The score builds on the regularisation applied throughout the AAE training process, namely the reconstruction error loss, denoted by Eq. (\ref{equ:reconstruction_loss}), and the adversarial loss, denoted by Eq. (\ref{equ:adversarial_loss}), described in the following. 

The reconstruction loss promotes the AAE to learn a set of non-overlapping latent journal entry representation. However, this may result in a highly "fractured" latent space in which deviating representations are learned for similar journal entries. The additionally applied adversarial loss prevents the fracturing problem. It forces the learned representations $z$ to reside within the high probability density regions of the imposed prior distribution $p(z)$. To partition the latent space into semantic regions we impose a multi-modal prior, e.g., a mixture of Gaussians. Thereby, the interaction of both regularising losses forces the AAE to learn groups of semantic similar journal entries located in close spatial proximity to the modes of the imposed prior.

As a result, the AAE learns with progressing training, a model that disentangles the underlying generative processes of journal entries in the latent space $z$. Each group of learned representations corresponds to a distinct generative process of journal entries, e.g., depreciation postings or vendor payment postings. To detect potential accounting anomalies, we investigate the individual entries of each group in terms of potential "violations" of one of the two applied regularising losses. We hypothesize, that anomalous journal entries can be captured by either (1) their latent divergence from the modes of the imposed prior or (2) an increased reconstruction error. Thereby, the type of violation also reveals the anomaly class of the investigated entry, as described in the following:

\textbf{Mode Divergence (MD)}: Journal entries that exhibit anomalous attribute values (global anomalies) result in an increased divergence from the imposed multi-modal prior, e.g., as in this work the divergence to the modes of an imposed mixture of multivariate isotropic Gaussians $\mathcal{N}(\mu,\mathcal{I})$, where $\mu \in \mathcal{R}^m$ defines the $\tau$ modes of the distinct Gaussians denoted by $\mu =\{\mu^1 \ldots \mu^\tau \}$. Throughout the AAE training, the entries will be "pushed" towards the high probability density regions of the prior by the regularization. In order to be able to discriminate between the imposed prior and the learned aggregated posterior the AAE aims to keep the majority of the entries within the high-density regions (modes) of the prior. In contrast, representations that correspond to rare or anomalous journal entries will tend to differ from the imposed modes and be placed in the priors low-density regions. We use this characteristic and obtain an entry's $x^{i}$ mode divergence $D$ as the Euclidean distance of the entry's learned representation $z^i$ to its closest mode $\mu^\tau$. Formally, we derive the mode divergence as denoted by $D_{\theta^*}^{\tau}(z^{i};\mu) = \min\limits_{\tau} \lVert z^i-\mu^\tau \rVert^2 $ under optimal model parameters $\theta^*$. Finally, we calculate the normalized mode divergence $MD$ as expressed by:

\begin{equation}
MD_{\theta^*}^{\tau}(x^{i}) = \frac{D_{\theta^*}^{\tau}(z^i;\mu) - D_{\theta^*, min}^{\tau}}{D_{\theta^*, max}^{\tau} - D_{\theta^*, min}^{\tau}},
\end{equation}

\noindent where $D_{min}$ and $D_{max}$ denotes the min- and max-values of the obtained mode divergences given by $D_{\theta^*}$ and closest mode $\tau$. 

\textbf{Reconstruction Error (RE)}: Journal entries that exhibit anomalous attribute value co-occurrences (local anomalies) tend to result in an increased reconstruction error \cite{schreyer2017}. This is caused by the compression capability of the AAE architecture. Anomalous and therefore unique attribute co-occurrences exhibit an increased probability of getting lost in the encoders "lossy" compression. As a result, their low dimensional representation will overlap with regular entries in the latent space and are not reconstructed correctly by the decoder. Formally, we obtain the reconstruction error $E$ of each entry $x^i$ and its reconstruction $\hat{x}^i$ as the squared-difference denoted by $E_{\theta^*}^{\tau}(x^{i};\hat{x}^{i}) = \frac{1}{k} \sum_{j=1}^{k}{(x^{i}_{j} - \hat{x}^{i}_{j})}^2$ under optimal model parameters $\theta^*$. Finally, we calculate the normalized reconstruction error $RE$ as expressed by:

\begin{equation}
RE_{\theta^*}^{\tau}(x^{i};\hat{x}^{i}) = \frac{E_{\theta^*}^{\tau}(x^i;\hat{x}^{i}) - E_{\theta^*, min}^{\tau}}{E_{\theta^*, max}^{\tau} - E_{\theta^*, min}^{\tau}},
\end{equation}

\noindent where $E_{min}$ and $E_{max}$ denotes the min- and max-values of the obtained reconstruction errors given by $E_{\theta^*}$ and closest mode $\tau$.

\textbf{Anomaly Score (AS)}: Quantifying both characteristics for a given journal entry, we can reasonably conclude (1) if the entry is anomalous and (2) if it was created by a "regular" business activity. To detect global and local accounting anomalies in real-world audit scenarios we propose to score each journal entry $x^i$ by its normalized reconstruction error $RE$ regularized and normalized mode divergence $MD$ given by:

\begin{equation}
AS^{\tau}(x^{i};\hat{x}^{i}) = \alpha \times RE_{\theta^*}^{\tau}(x^{i};\hat{x}^{i}) + (1-\alpha) \times MD_{\theta^*}^{\tau}(x^{i}),
\end{equation} 

\noindent for each individual journal entry $x^{i}$ and optimal model parameters $\theta^*$ and closest mode $\tau$. We introduce $\alpha$ as a factor to balance both characteristics. 


\section{Experimental Setup}
\label{sec:experiments}

In this section, we describe the experimental setup and model training. We evaluate the anomaly detection performance of the proposed scoring based on two datasets of journal entries.

\begin{figure*}[ht!]
    \begin{center}
        \includegraphics[width=0.28\textwidth]{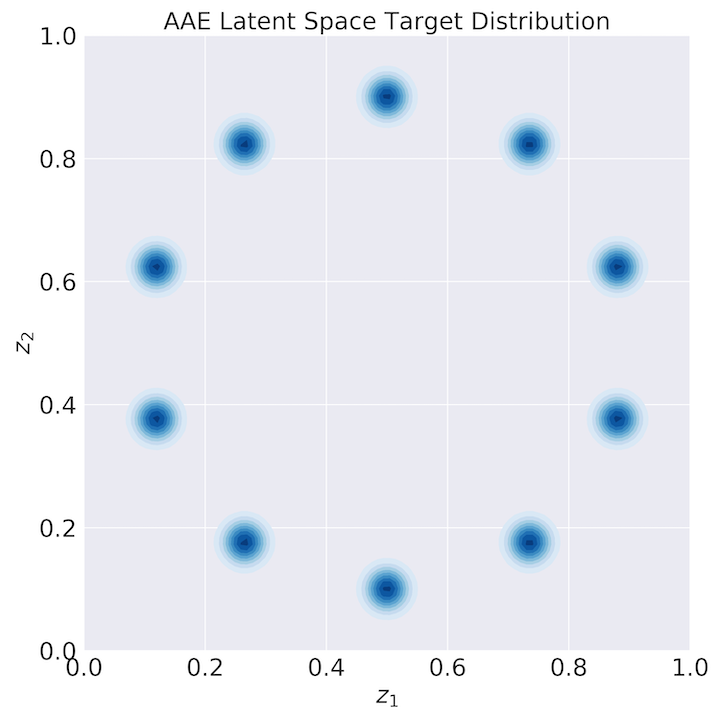}
        \includegraphics[width=0.28\textwidth]{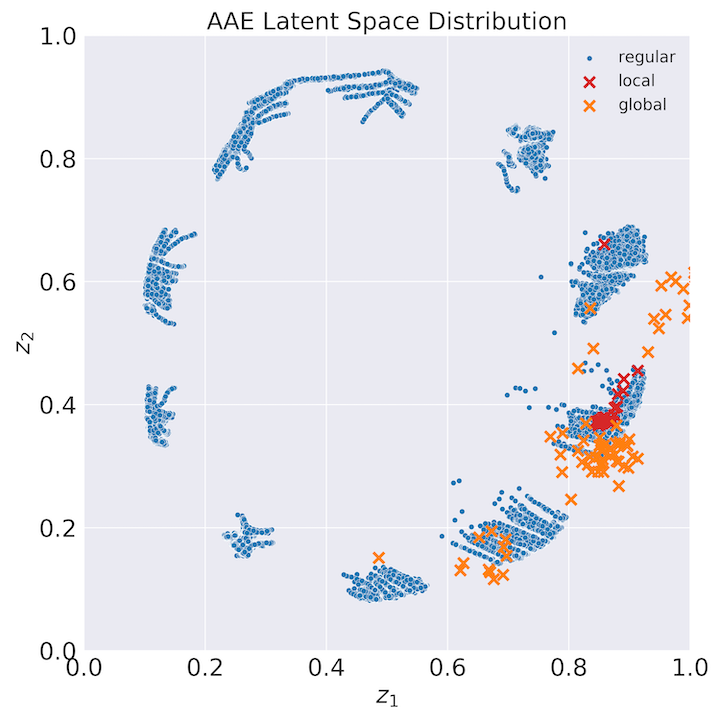}
         \includegraphics[width=0.28\textwidth]{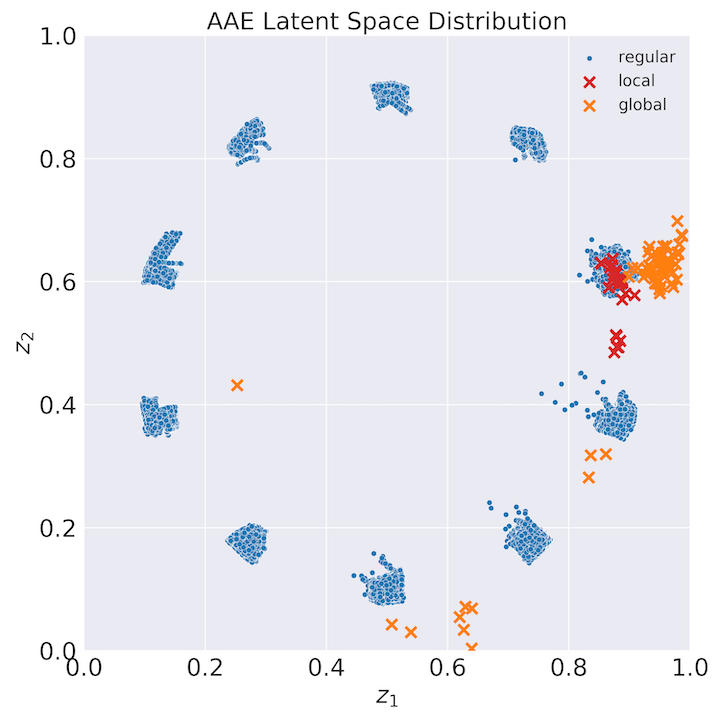}
    \end{center}
    \caption{Exemplary AAE latent space distribution of dataset B with progressing network training: imposed prior distribution $p(z)$ consisting of a mixture of $\tau=10$ Gaussians (left), learned aggregated posterior distribution $g_{\theta}(z|x)$ after 100 training epochs (center), learned aggregated posterior distribution $g_{\theta}(z|x)$ after 2,000 training epochs (right).}
    \label{fig:training_process}
\end{figure*}

\subsection{Datasets and Data Preparation}

In general, SAP ERP systems record journal entries and their corresponding attributes predominantly in two database tables: (1) the table "Accounting Document Headers" (technically: "BKPF") contains the meta-information of a journal entry, such as document id, type, date, time, or currency, while (2) the table  "Accounting Document Segments" (technically: "BSEG") contains the entry details, such as posting key, general ledger account, debit-credit information, or posting amount. In the context of this work, we extract a subset of the most discriminative journal entry attributes of the "BKPF" and "BSEG" table.

In our experiments we use two datasets of journal entries: a real-world and a synthetic dataset referred to as \textit{dataset A} and \textit{dataset B} in the following. Dataset A is an extract of an SAP ERP instance and encompasses the entire population of journal entries of a single fiscal year\footnote{In compliance with strict data privacy regulations, all journal entry attributes of dataset A have been anonymized using an irreversible one-way hash function during the data extraction process. To ensure data completeness, the journal entry based general ledger balances were reconciled against the standard SAP trial balance reports e.g. the SAP 'RFBILA00' report.}. Dataset B is an excerpt of the synthetic dataset presented in \cite{paysim}\footnote{The original dataset is publicly available via the Kaggle predictive modeling and analytics competitions platform and can be obtained using the following link: https://www.kaggle.com/ntnu-testimon/paysim1.}. The majority of attributes recorded in ERP systems correspond to categorical (discrete) variables, e.g. posting date, account, posting type, currency. We pre-process the categorical journal entry attributes to obtain a binary ("one-hot" encoded) representation of each journal entry. 

To allow for a detailed analysis and quantitative evaluation of the experiments we inject a small fraction of synthetic global and local anomalies into both datasets. Similar to real audit scenarios this results in a highly unbalanced class distribution of anomalous vs. regular day-to-day entries. The injected global anomalies consist of attribute values not evident in the original data while the local anomalies exhibit combinations of attribute value subsets not occurring in the original data. The true labels are available for both datasets. Each journal entry is labeled as either (1) synthetic \textit{global anomaly}, (2) synthetic \textit{local anomaly} or (3) non-synthetic \textit{regular entry}. The following descriptive statistics summarize both datasets:

\begin{itemize}
\item \textbf{Dataset A:} contains a total of $307,457$ journal entry line items comprised of six categorical and two numerical attributes. The encoding resulted in a total of $401$ encoded dimensions for each entry $x^{i} \in \mathcal{R}^{401}$; and, In total $95$ ($0.03\%$) synthetic anomalous journal entries have been injected into the dataset. These entries encompass $55$ ($0.016\%$) global anomalies and $40$ ($0.015\%$) local anomalies.
\item \textbf{Dataset B:} contains a total of $533,009$ journal entry line items comprised of six categorical and two numerical attributes. The encoding resulted in a total of $618$ encoded dimensions for each entry $x^{i} \in \mathcal{R}^{618}$. In total $100$ ($0.018\%$) synthetic, anomalous journal entries have been injected into the dataset. These entries encompass $70$ ($0.013\%$) global anomalies and $30$ ($0.005\%$) local anomalies.
\end{itemize}

Figure \ref{fig:input_data_distribution} illustrates an exemplary distribution of the attributes primarily investigated during and audit, namely the 'account key' (technically: 'KTOSL') attribute values, the log-normalized 'local' and 'foreign currency amount' (technically: 'DMBTR' and 'WRBTR') attribute values, as well as the 'posting key' (technically: 'BSCHL') and 'general ledger account' (technically: 'HKONT') observable in Dataset B.

\subsection{Adversarial Autoencoder Training}

Our architectural setup follows the AAE architecture \cite{makhzani2015} as shown in Fig. \ref{fig:architecture}, comprised of three distinct neural networks that are trained in parallel. The encoder network $q_{\theta}$ uses Leaky Rectified Linear Unit (LReLU) activation functions \cite{xu2015} except in the last "bottleneck" layer. Both the decoder network $p_{\theta}$ and the discriminator $d_{\phi}$ network use LReLUs in all layers except the output layers where a Sigmoid activation function is used. Table \ref{tab:architecture} depicts the architectural details of the networks which are implemented using PyTorch \cite{paszke2017}.

Training stability is a main challenge in adversarial training \cite{arjovsky2017} and we face a variety of collapsing and non-convergence scenarios. To determine a stable training setup we sweep the learning rates of the encoder and decoder networks through the interval $\eta \in [10^{-05}, 10^{-02}]$, and the learning rates of the discriminator network through the interval $\eta \in [10^{-07}, 10^{-03}]$. Ultimately, we use the following constant learning rates to learn a stable model of each dataset:

\begin{itemize}
\item \textbf{Dataset A:} $\eta = 10^{-4}$ for the encoder and the decoder network, $\eta = 10^{-5}$ for the discriminator network; and,  
\item \textbf{Dataset B:} $\eta = 10^{-3}$ for the encoder and the decoder network, $\eta = 10^{-5}$ for the discriminator network.
\end{itemize}

 \begin{table}[ht!]
  \caption{Neurons per layer $\ell$ of the distinct networks that comprise the AAE architecture \cite{makhzani2015}: encoder $q_{\theta}$, decoder $p_{\theta}$ and discriminator $d_{\phi}$ neural network.} 
  \fontsize{8}{6}\selectfont
  \centering
  \begin{tabular}{l c | c c c c c c c c }
    \toprule
        \multicolumn{1}{l}{Net}
        & \multicolumn{1}{c}{Dataset}
        & \multicolumn{1}{c}{$\ell$ = 1}
        & \multicolumn{1}{c}{2}
        & \multicolumn{1}{c}{3}
        & \multicolumn{1}{c}{4}
        & \multicolumn{1}{c}{5}
        & \multicolumn{1}{c}{6}
        & \multicolumn{1}{c}{7}
        & \multicolumn{1}{c}{8}
        \\
    \midrule
    $q_{\theta}(z|x)$ & A & 256 & 128 & 64 & 32 & 16 & 8 & 4 & 2 \\
    $p_{\theta}(\hat{x}|z)$ & A & 2 & 4 & 8 & 16 & 32 & 64 & 128 & 256 \\
    $d_{\phi}(z)$ & A & 128 & 64 & 32 & 16 &  &  &  & \\
    \midrule
    $q_{\theta}(z|x)$ & B & 256 & 64 & 16 & 4 & 2 & - & - & - \\
    $p_{\theta}(\hat{x}|z)$ & B & 2 & 4 & 16 & 64 & 256 & - & - & - \\
    $d_{\phi}(z)$ & B & 256 & 64 & 16 & 4 & 1 & - & - & - \\
    \bottomrule \\
  \end{tabular}
    \label{tab:architecture}
 \end{table} 
 
We train the AAE with mini-batch wise SGD for max. 10,000 training epochs and apply early stopping once the reconstruction loss converges. In accordance with \cite{xu2015} we set the scaling factor of the LReLUs to $\alpha = 0.4$ and initialized the AAE parameters as described in \cite{glorot2010}. A mini-batch size of 128 journal entries is used in both the reconstruction and the regularization phase. We use Adam optimization \cite{kingma2014} and set $\beta_{1}=0.9$ and $\beta_{2}=0.999$ in the optimization of the network parameters. In the reconstruction phase, we use a combined loss function $\mathcal{L}_{\theta}$ to optimize the encoder $q_{\theta}$ and decoder net $p_{\theta}$ parameters. For each journal entry we calculate (1) the cross-entropy reconstruction error $\mathcal{L}^{CE}_{\theta}$ of the categorical attribute value encodings $x^{i}_{cat}$, e.g., the encoded general ledger account id, and (2) the mean-squared reconstruction error $\mathcal{L}^{MSE}_{\theta}$ of the numerical attribute value encodings $x^{i}_{con}$, e.g., the encoded posting amount, formally expressed by:  

\begin{equation}
    \mathcal{L}_{\theta}(x^{i};\hat{x}^{i}) = \gamma \hspace{1mm} \mathcal{L}^{CE}_{\theta}(x^{i}_{cat};\hat{x}^{i}_{cat}) + (1 - \gamma) \hspace{1mm} \mathcal{L}^{MSE}_{\theta}(x^{i}_{con};\hat{x}^{i}_{con})
    \vspace{2mm}
    \label{equ:reconstruction_loss_details}
\end{equation}

\noindent were the parameter $\gamma$ balances both losses. In this initial work, we set $\gamma = \frac{2}{3}$ in all our experiments to account for the higher amount of categorical attributes in both datasets. In the regularization phase, we calculate the adversarial loss, according to equation \ref{equ:adversarial_loss}, when optimizing the parameters of the discriminator $d_{\phi}$.

To partition the learned journal entry representations, we sample from a prior distribution $p(z)$ comprised of a mixture of $\tau$ multivariate isotropic Gaussians $\mathcal{N}(\mu,\mathcal{I})$, where $\mu \in \mathcal{R}^{2}$. Thereby, $\tau$ is a hyperparameter we evaluate when sampling of $\tau \in \{5, 10, 15\}$ Gaussians. Figure \ref{fig:training_process} shows an exemplary prior consisting of $\tau=10$ Gaussians as well as the learned aggregated posterior distributions after 100 and 2,000 training epochs.

\section{Experimental Results}
\label{sec:results}

In this section, we first assess the semantic partitioning of the journal entries by the imposed prior distributions. Afterward, we examine the anomalies detected of each semantic partition.

\begin{figure*}[ht!]
    \begin{center}
        \includegraphics[width=0.28\textwidth]{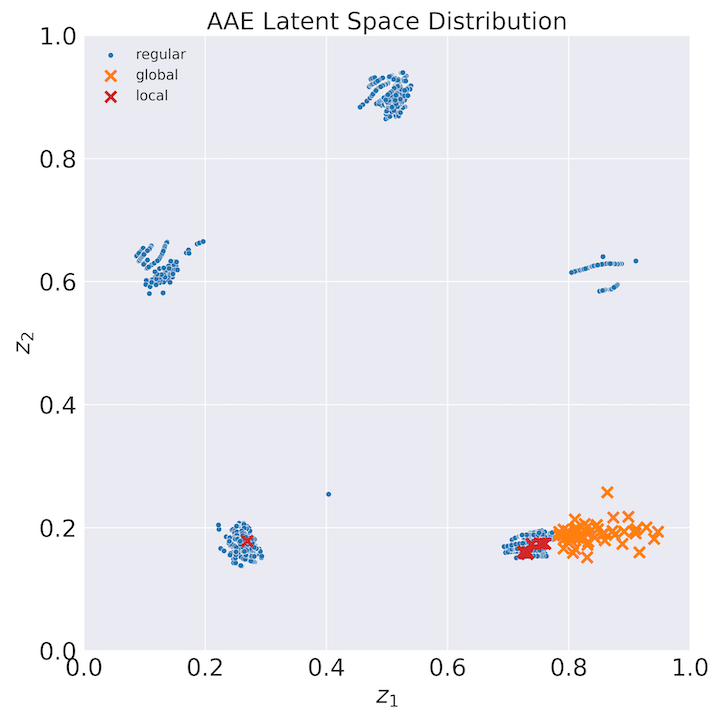}
        \includegraphics[width=0.28\textwidth]{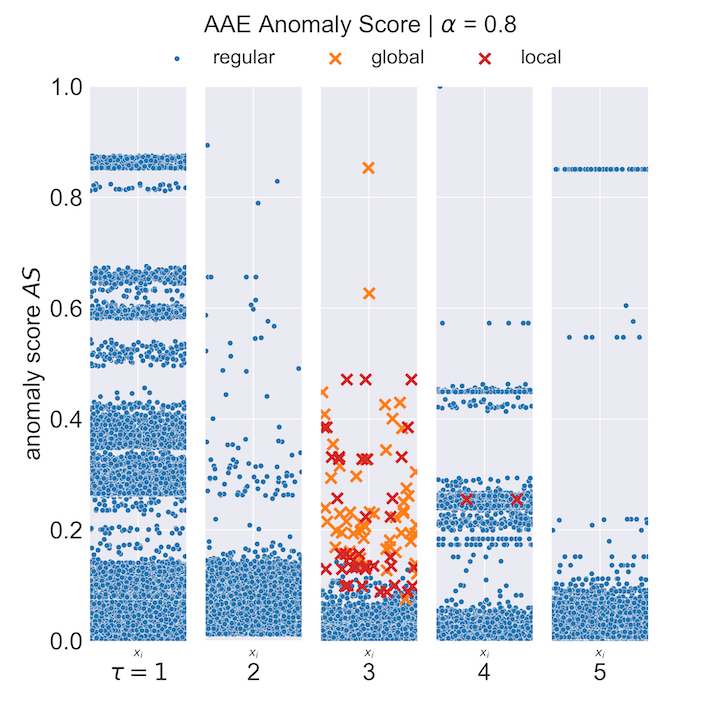}
        \includegraphics[width=0.30\textwidth]{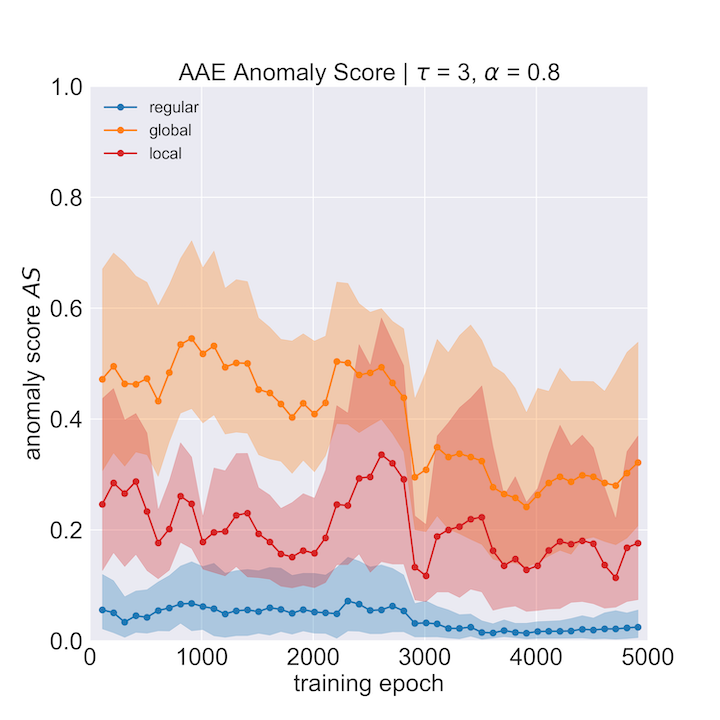}
    \end{center}
    \caption{Learned AAE latent space representations of the journal entries contained in dataset A after training the AAE for 5,000 epochs and imposing a mixture of $\tau=5$ Gaussians (left), the anomaly scores $AS$ obtained for $\alpha=0.8$ of each journal entry $x_{i}$ and corresponding mode $\mu_{\tau}$ (center), the anomaly score distribution (bold line defines the median, upper and lower bound define the $0.05$ and $0.95$ quantile of the distribution) obtained of each journal entry class with progressing network training (right).}
    \label{fig:latent_space_1}
\end{figure*}

\textbf{Semantic partitioning:} We qualitatively review the latent space partitioning of the journal entries and assess the accounting specific semantics that is learned by each mode. Figure \ref{fig:latent_space_1} shows the partitioning result of dataset A, where $\tau=5$ Gaussians (see appxs. for results of varying $\tau$ and dataset B). It can be observed that the AAE learned a rather clean separation of the regular journal entries. The review of the journal entries accounting specific semantics captured by each mode and dataset revealed:

\begin{itemize}
\item \textbf{Dataset A:} The entries of each partition exhibit a high semantic similarity while each partition corresponds to a general accounting process, such as (1) automated payment run entries postings, (2) outgoing customer invoices, and (3) material movements. 
\item \textbf{Dataset B:} Similarly, the entries of each partition exhibit a high semantic similarity and correspond to the following general accounting processes (1) foreign and domestic invoice postings, (2) purchase of goods, (3) manual payments.
\end{itemize}

The experimental results when imposing $\tau \in \{5, 10, 15\}$ Gaussians on the latent space of each dataset are presented in the appendix of this work. The results show that the AAE is capable of learning a semantic partition of a given set of journal entries according to that disentangles the entries underlying generative processes. The learned partition provides the auditor a holistic view on a given set of accounting data subject to audit. Furthermore, it allows to effectively obtain a representative and interpretable sample of the data and thereby reduces the audits sampling risk.

\begin{figure*}[ht!]
    \begin{center}
        \includegraphics[width=0.28\textwidth]{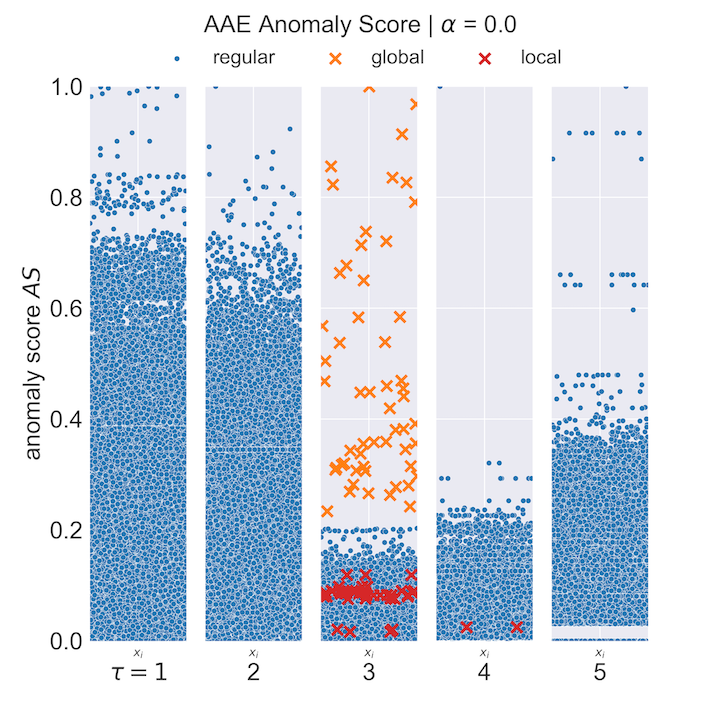}
        \includegraphics[width=0.28\textwidth]{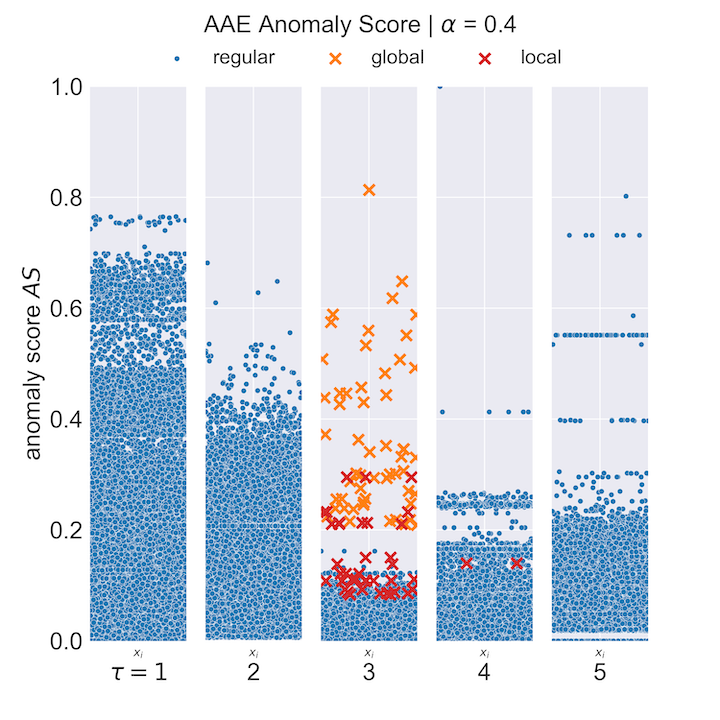}
         \includegraphics[width=0.28 \textwidth]{ep_5000_05_anomaly_score_alpha_0_8_strip.png}
    \end{center}
    \caption{Anomaly scores $AS$ obtained by the application of distinct $RE$ vs. $MD$ balance factors $\alpha$ after training the AAE for 5,000 epochs on dataset A (see appxs. for results of varying $\tau$ and dataset B) and imposing a mixture of $\tau=5$ Gaussians. It can be observed that decreasing $\alpha$ results in an improved detection of global anomalies (left). In contrast, increasing $\alpha$ results in an improved detection of local anomalies (right).}
    \label{fig:latent_space_2}
\end{figure*}

\begin{table}[ht!]
\caption{Mean anomaly score $AS$ obtained per journal entry class and $\alpha=0.8$ by imposing a prior distribution consisting of $\tau = 5$ ($10$, and $15$) mixture of Gaussians and training the AAE for 5,000 (10,000, and 15,000) epochs (variances originate from the distinct parameter initialization seeds).} 
\fontsize{8}{6}\selectfont
\centering
\begin{tabular}{l c | c c | c c | c c }
\toprule
    \multicolumn{1}{l}{Class}
    & \multicolumn{1}{c}{Data}
    & \multicolumn{2}{c}{$AS$, $\tau$ = 5}
    & \multicolumn{2}{c}{$AS$, $\tau$ = 10}
    & \multicolumn{2}{c}{$AS$, $\tau$ = 15}
    \\
\midrule
global & A & 0.295 & $\pm$ 0.233 & 0.448 & $\pm$ 0.207 & 0.532 & $\pm$ 0.244 \\
local  & A & 0.248 & $\pm$ 0.276 & 0.275 & $\pm$ 0.143 & 0.446 & $\pm$ 0.202 \\
regular & A & 0.045 & $\pm$ 0.076 & 0.053 & $\pm$ 0.085 & 0.110 & $\pm$ 0.034 \\
\midrule
global & B & 0.508 & $\pm$ 0.249 & 0.442 & $\pm$ 0.245 & 0.437 & $\pm$ 0.241 \\
local & B & 0.357 & $\pm$ 0.260 & 0.164 & $\pm$ 0.148 & 0.273 & $\pm$ 0.228 \\
regular & B & 0.046 & $\pm$ 0.061 & 0.070 & $\pm$ 0.041 & 0.028 & $\pm$ 0.029 \\
\bottomrule \\
\end{tabular}
\label{tab:partition_scores}
\end{table}

\textbf{Anomaly detection:} In addition, we analyze the anomaly detection capability of the proposed anomaly score. Table \ref{tab:partition_scores} depicts the mean anomaly score $AS$ obtained for each journal entry class by imposing a prior distribution consisting of $\tau \in \{5, 10, 15\}$ mixture of Gaussians and training the AAE for 5,000 epochs. The quantitative results show the distinct journal entry classes (global, local, and regular entries) can be distinguished according to their anomaly score in both datasets. Figure \ref{fig:latent_space_1} exemplary shows the anomaly scores obtained for dataset A (see appxs. for results of dataset B) of each journal entry and corresponding partition $\tau$ as well as the distribution of the obtained individual anomaly scores. Figure \ref{fig:latent_space_2} illustrates the change in anomaly scoring when varying the $\alpha$ parameter of the $AS$. It can be observed that increasing $\alpha$ (and therefore the weight on the reconstruction error of the score) improves the ability to detect local accounting anomalies in the dataset. 

We also qualitatively evaluate the characteristics of the anomalies detected in each partition. Therefore, we review journal entries that correspond to a high anomaly score but have not been synthetically injected as anomalies into the evaluated datasets. Thereby, we interpret the detected anomalies of each mode $\tau$ in the context of the modes regular entries:

\begin{itemize}
\item \textbf{Global anomalies} exhibit a low semantic similarity to the regular entries of a mode. The detected entries correspond to rarely observable attribute values and accounting "exceptions"', e.g., unusual purchase order amounts or high depreciation, year-end as well as impairment postings.
\item \textbf{Local anomalies} exhibit high semantic similarity to the regular entries of a mode. The detected entries correspond to rarely observable attribute value combinations, e.g., system users that switched departments, postings exhibiting unusual general ledger account combinations. 
\end{itemize}

In summary, these results lead us to conclude that the proposed anomaly score can be utilized as a highly adaptive anomaly assessment of financial accounting data. It furthermore provides the ability to interpret the detected anomalies of a particular mode in the context of the modes regular journal entry semantics. Initial feedback received by auditors on the detected anomalies underpinned not only their relevance from an accounting perspective.
 
\section{Summary}
\label{sec:conclusion}

In this work, we showed that Adversarial Autoencoder (AAE) neural networks can be trained to learn a semantic meaningful representation of journal entries recorded in real-world ERP systems. We also provided initial evidence that such representations provide a holistic view of the entries and disentangle the underlying generative processes. We believe that the presented approach enables a human auditor or forensic accountant with the ability to sample journal entries for a detailed audit in an interpretable manner and therefore reduce the "sampling risk". In addition, we proposed a novel anomaly score that combines and entry's learned representation and reconstruction error. We demonstrated that the scoring can be interpreted as a highly adaptive and unsupervised anomaly assessment to detect global and accounting anomalies.

We plan to conduct a more detailed investigation of the journal entries' latent space disentanglement. Given the tremendous amount of journal entries annually recorded by organizations, an automated semantic disentanglement improves the transparency of entries to be audited and can save auditors considerable time.

\begin{acks}
We thank the members of the statistics department at Deutsche Bundesbank and PwC Europe's Forensic Services for their valuable review and remarks. Opinions expressed in this work are solely those of the authors, and do not necessarily reflect the view of the Deutsche Bundesbank or PricewaterhouseCoopers (PwC) International Ltd. and its network firms.\\ 

\end{acks}

\bibliographystyle{ACM-Reference-Format}
\bibliography{kdd2019references}

\appendix
\onecolumn
\section*{Appendix}

\subsection*{Experimental Results - Dataset A}

\begin{figure*}[ht!]
    \begin{center}
        \includegraphics[width=0.28\textwidth]{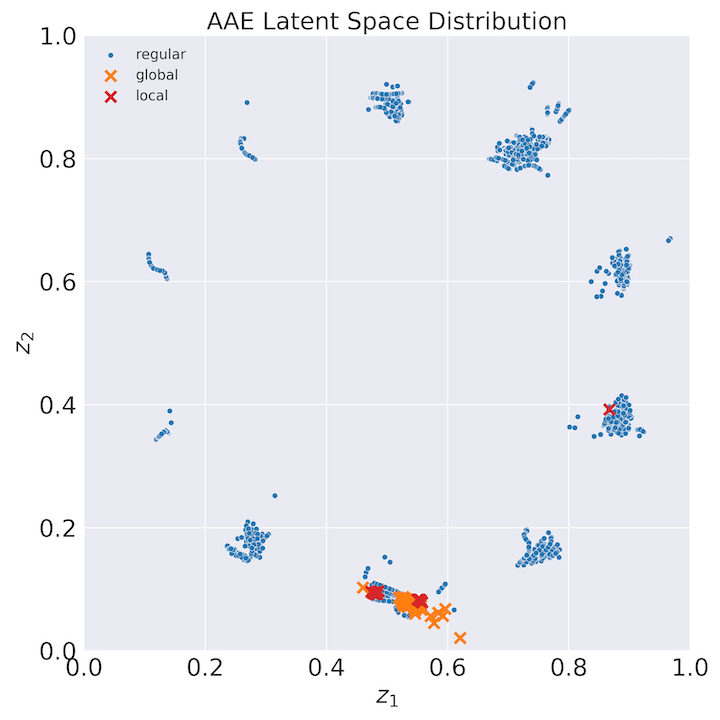}
        \includegraphics[width=0.28\textwidth]{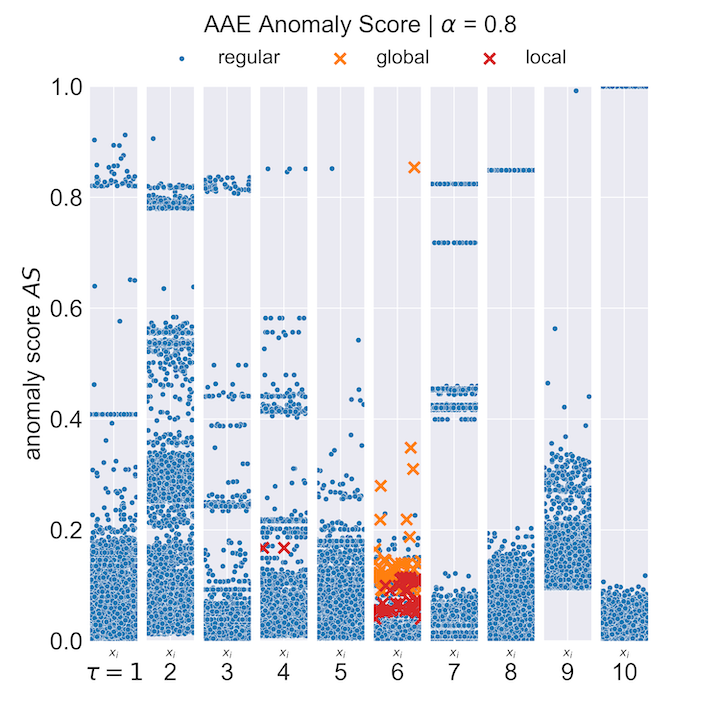}
        \includegraphics[width=0.30\textwidth]{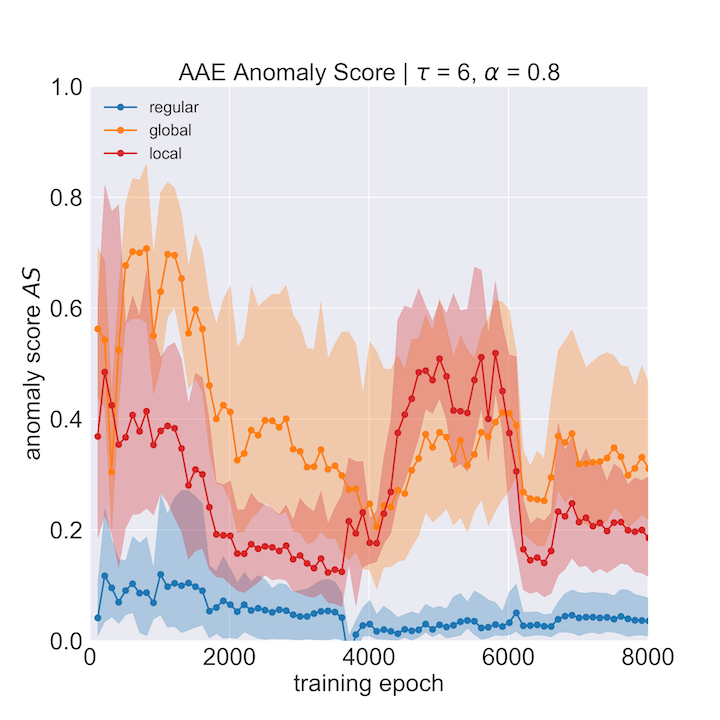}
    \end{center}
    \caption{Learned AAE latent space representations of the journal entries contained in dataset A after training the AAE for 10,000 epochs and imposing a mixture of $\tau=10$ Gaussians (left), the anomaly scores $AS$ obtained for $\alpha=0.8$ of each journal entry $x_{i}$ and corresponding mode $\mu_{\tau}$ (center), the anomaly score distribution (bold line defines the median, upper and lower bound define the $0.05$ and $0.95$ quantile of the distribution) obtained of each journal entry class with progressing training (right).}
    \label{fig:latent_space_app_2}
\end{figure*}

\begin{figure*}[ht!]
    \begin{center}
        \includegraphics[width=0.28\textwidth]{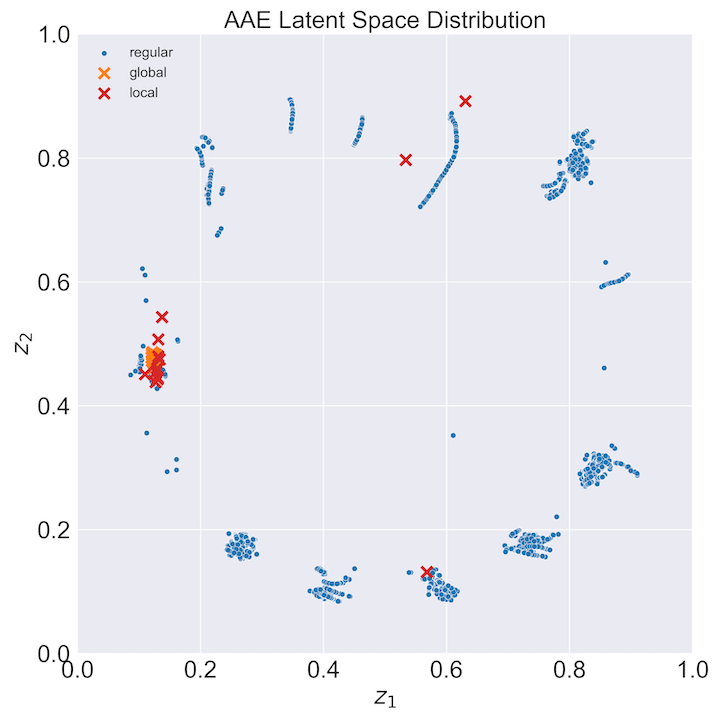}
        \includegraphics[width=0.28\textwidth]{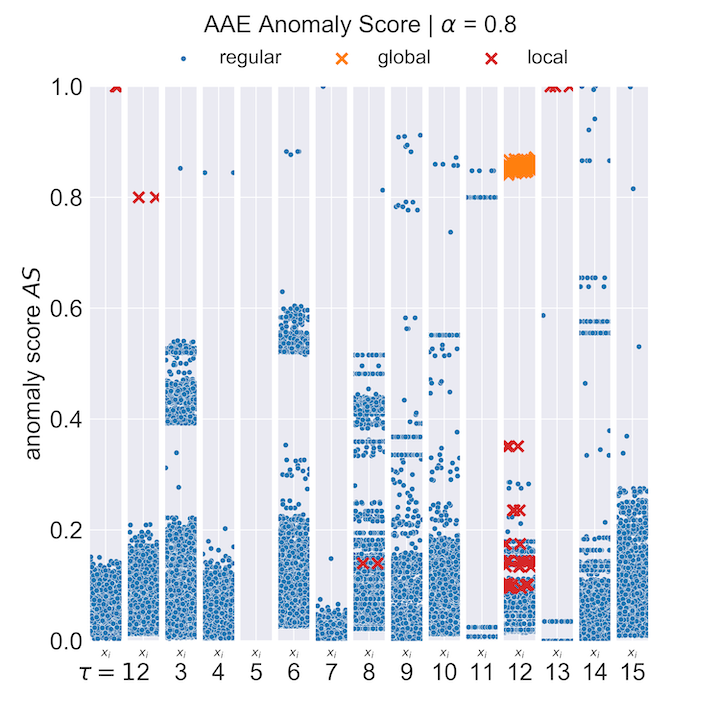}
        \includegraphics[width=0.30\textwidth]{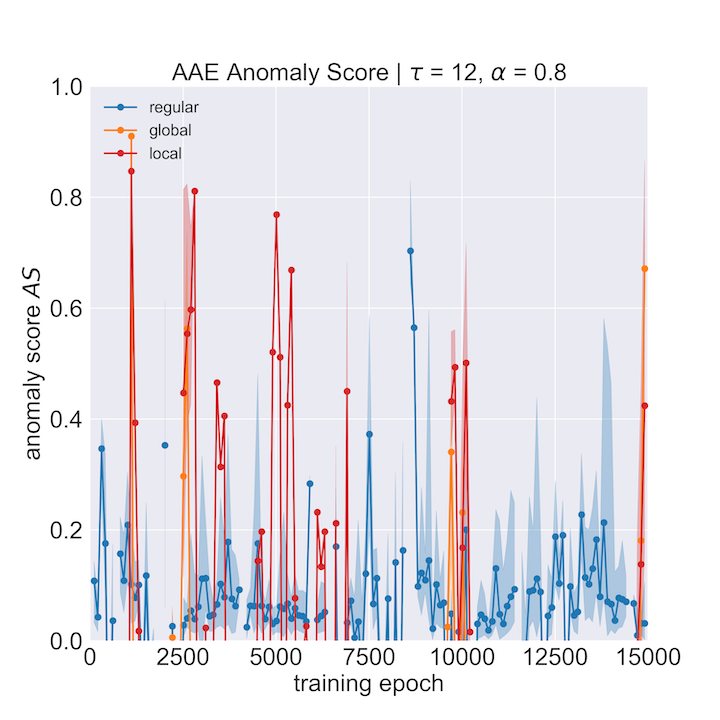}
    \end{center}
    \caption{Learned AAE latent space representations of the journal entries contained in dataset A after training the AAE for 15,000 epochs and imposing a mixture of $\tau=15$ Gaussians (left), the anomaly scores $AS$ obtained for $\alpha=0.8$ of each journal entry $x_{i}$ and corresponding mode $\mu_{\tau}$ (center), the anomaly score distribution (bold line defines the median, upper and lower bound define the $0.05$ and $0.95$ quantile of the distribution) obtained of each journal entry class with progressing training (right). It can be observed, based on the anomaly scores obtained with progressing training of mode $\tau=12$, that mode stability wasn't reached entirely after training the AAE for 15,000 epochs.}
    \label{fig:latent_space_app_3}
\end{figure*}

\newpage

\subsection*{Experimental Results - Dataset B}

\begin{figure*}[ht!]
    \begin{center}
        \includegraphics[width=0.28\textwidth]{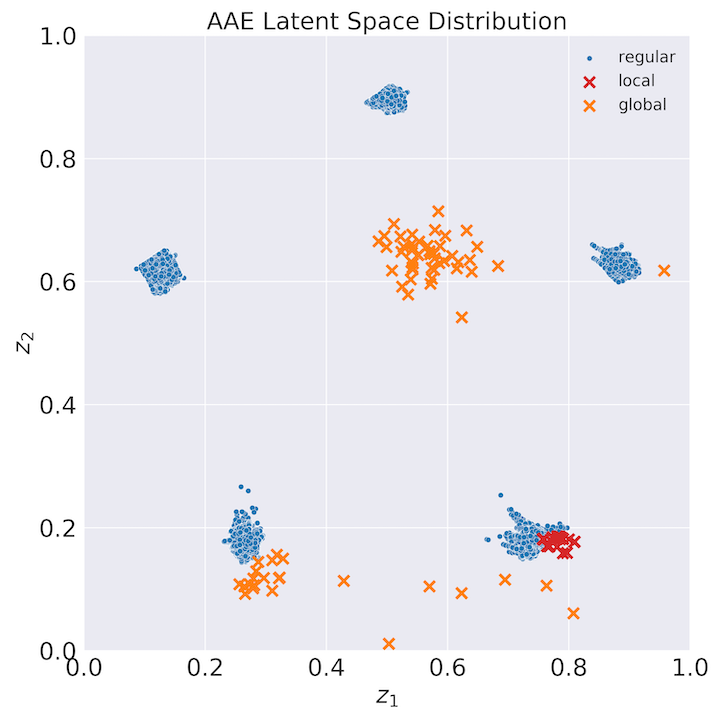}
        \includegraphics[width=0.28\textwidth]{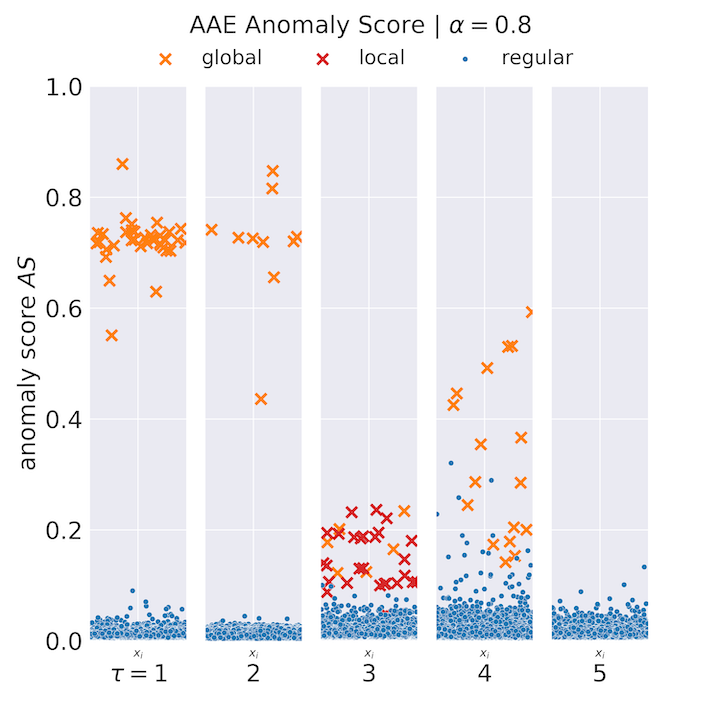}
        \includegraphics[width=0.30\textwidth]{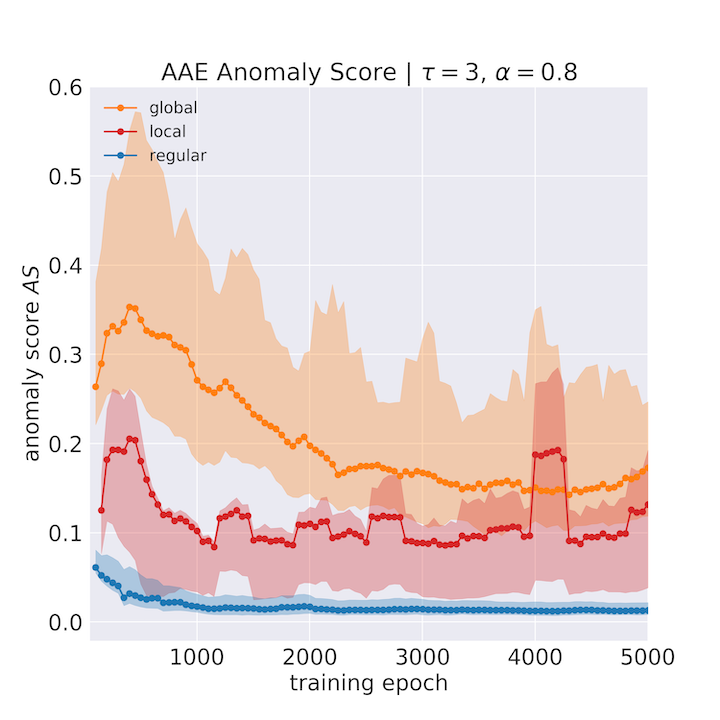}
    \end{center}
    \label{fig:latent_space_app_4}
\end{figure*}

\begin{figure*}[ht!]
    \begin{center}
        \includegraphics[width=0.28\textwidth]{gs_10_latent_space_ep_5000_gs_10.png}
        \includegraphics[width=0.28\textwidth]{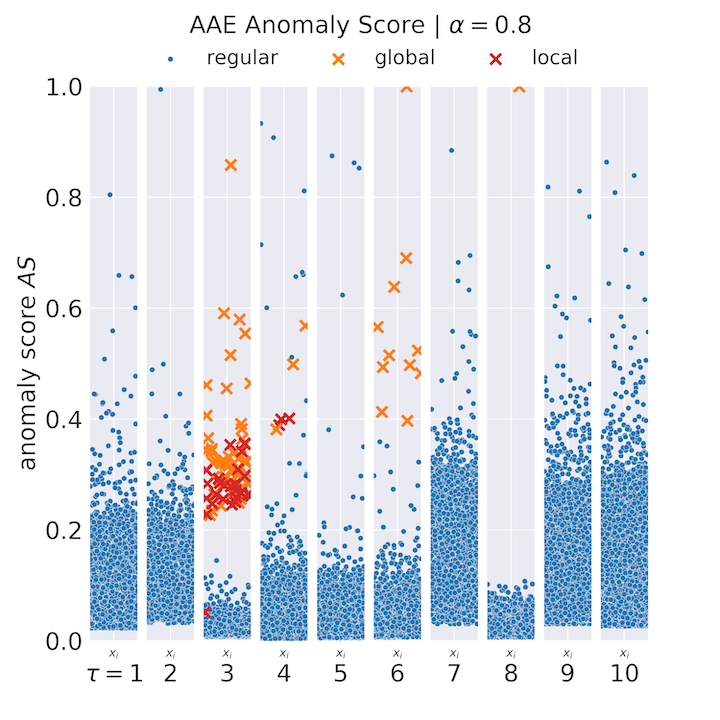}
        \includegraphics[width=0.30\textwidth]{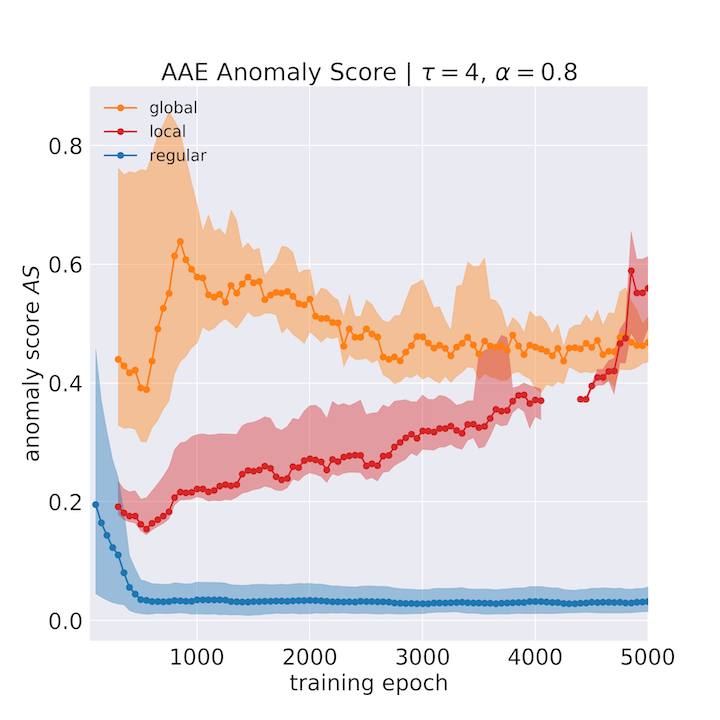}
    \end{center}
    \label{fig:latent_space_app_5}
\end{figure*}

\begin{figure*}[ht!]
    \begin{center}
        \includegraphics[width=0.28\textwidth]{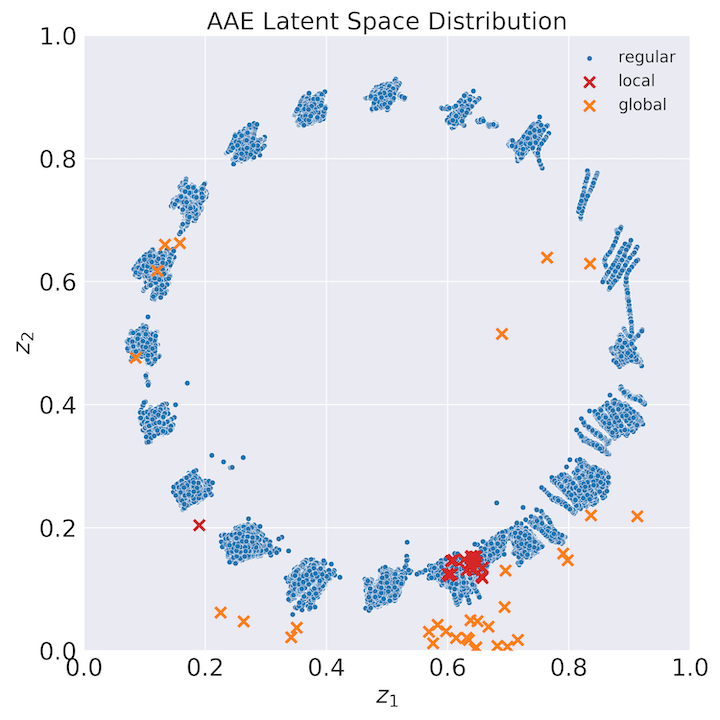}
        \includegraphics[width=0.38\textwidth]{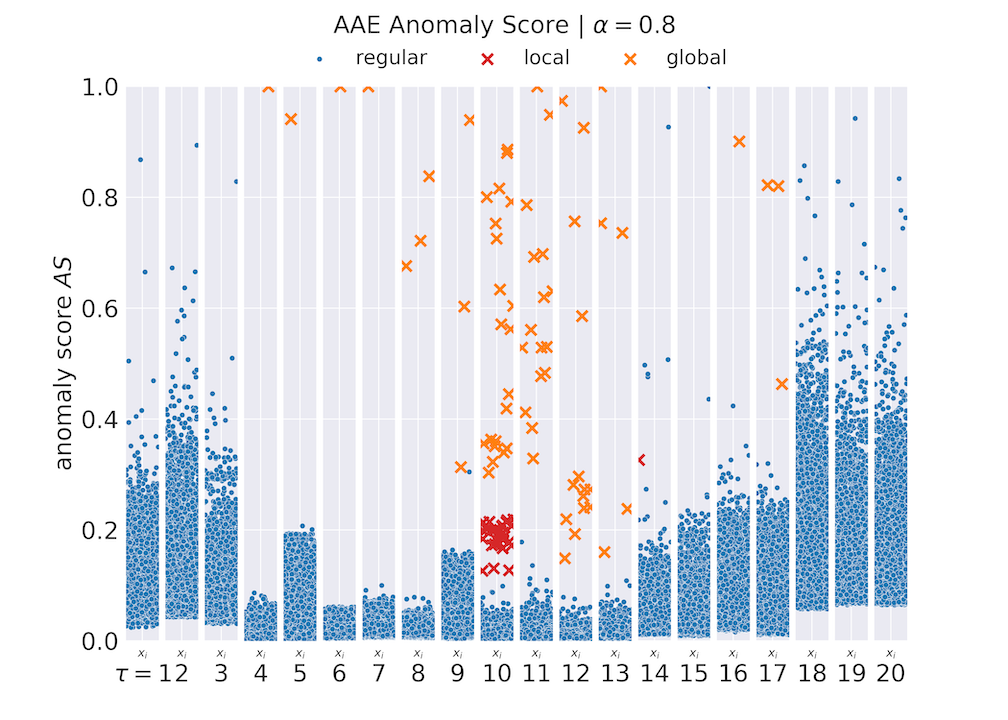}
        \includegraphics[width=0.30\textwidth]{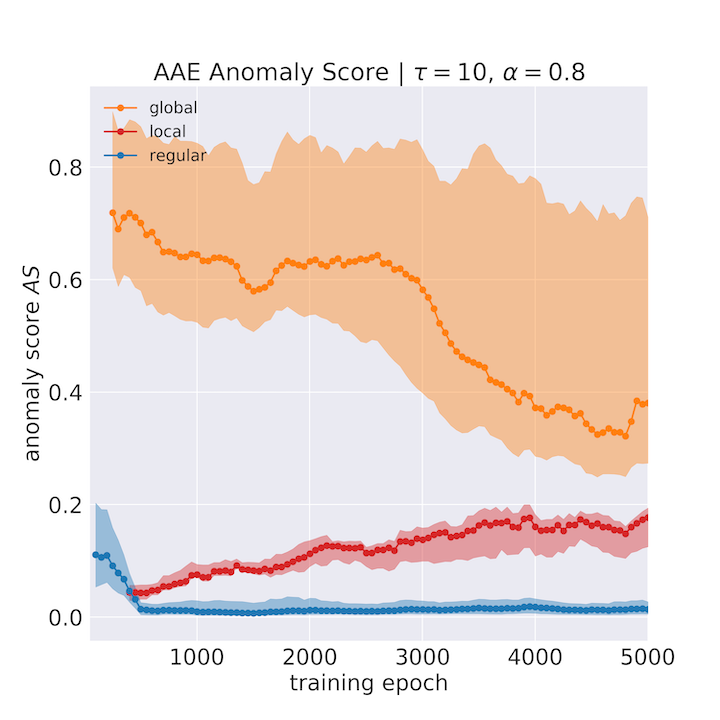}
    \end{center}
    \caption{Learned AAE latent space representations of the journal entries contained in dataset B after training the AAE for 5,000 epochs and imposing a mixture of $\tau=5, 10$ and $20$ Gaussians (top to bottom) (left), the anomaly scores $AS$ obtained for $\alpha=0.8$ of each journal entry $x_{i}$ and corresponding mode $\mu_{\tau}$ (center), the anomaly score distribution (bold line defines the median, upper and lower bound define the $0.05$ and $0.95$ quantile of the distribution) obtained of each journal entry class with progressing training (right).}
    \label{fig:latent_space_app_6}
\end{figure*}

\end{document}